\documentclass{article}
\pdfoutput=1

\usepackage[nonatbib, final]{neurips_2019}

\usepackage[numbers]{natbib}

\usepackage[utf8]{inputenc} % allow utf-8 input
\usepackage[T1]{fontenc}    % use 8-bit T1 fonts
\usepackage{hyperref}       % hyperlinks
\usepackage{url}            % simple URL typesetting
\usepackage{booktabs}       % professional-quality tables
\usepackage{amsfonts}       % blackboard math symbols
\usepackage{nicefrac}       % compact symbols for 1/2, etc.
\usepackage{microtype}      % microtypography
\usepackage{natbib}
\usepackage{graphicx}

\usepackage[utf8]{inputenc} % allow utf-8 input
\usepackage[T1]{fontenc}    % use 8-bit T1 fonts
\usepackage{hyperref}       % hyperlinks
\usepackage{url}            % simple URL typesetting
\usepackage{booktabs}       % professional-quality tables
\usepackage{amsfonts}       % blackboard math symbols
\usepackage{nicefrac}       % compact symbols for 1/2, etc.
\usepackage{microtype}      % microtypography
\usepackage{natbib}
\usepackage{graphicx}
\usepackage{rotating}
\usepackage{graphbox}
\usepackage{tabularx}

\usepackage{amsmath}
\usepackage{amssymb}

\newcommand{\todo}[1]{{\textbf{ [TODO: #1] }}}
\newcommand{\MA}[1]{{\textbf{ [MA: #1] }}}
\newcommand{\PS}[1]{{\textbf{ [PS: #1] }}}
\newcommand{\real}{\mathbb{R}}

\renewcommand{\todo}[1]{}
\renewcommand{\MA}[1]{}
\renewcommand{\PS}[1]{}

\title{Computational Mirrors: Blind Inverse Light Transport by Deep Matrix Factorization}

\author{%
    Miika Aittala \\ %\thanks{The source code and a supplemental video demonstrating the results can be found at the project webpage at \url{compmirrors.csail.mit.edu}.} \\
    MIT \\
    \texttt{miika@csail.mit.edu} \\
    \And
    Prafull Sharma \\
    MIT \\
    \texttt{prafull@mit.edu} \\
    \And
    Lukas Murmann \\
    MIT \\
    \texttt{lmurmann@mit.edu}\\
    \And
    Adam B. Yedidia \\
    MIT \\
    \texttt{adamy@mit.edu} \\
    \And
    Gregory W. Wornell \\
    MIT \\
    \texttt{gww@mit.edu} \\
    \And
    William T. Freeman \\
    MIT, Google Research \\
    \texttt{billf@mit.edu} \\
    \And
    Fr\'edo Durand \\
    MIT \\
    \texttt{fredo@mit.edu} \\
}

\begin{document}

\maketitle

\begin{abstract}
%\todo{whoever reads this, please volunteer to rewrite} We seek to recover a video of the motion taking place in a hidden region of a scene from the observation of the temporal changes of a natural complex and uncalibrated visible region. We solve the problem by factorizing the observed video into a matrix product between the hidden scene video and an unknown light transport matrix. This task is extremely ill-posed, as any non-negative factorization satisfies the data. Inspired by recent work on the Deep Image Prior, we parameterize the factor matrices using randomly initialized convolutional neural networks, and show that this factorization results in decompositions that have the desired appearance of natural images and videos.

We recover a video of the motion taking place in a hidden scene by observing changes in indirect illumination in a nearby uncalibrated visible region. We solve this problem by factoring the observed video into a matrix product between the unknown hidden scene video and an unknown light transport matrix. This task is extremely ill-posed, as any non-negative factorization will satisfy the data. Inspired by recent work on the Deep Image Prior, we parameterize the factor matrices using randomly initialized convolutional neural networks trained in a one-off manner, and show that this results in decompositions that reflect the true motion in the hidden scene.

\end{abstract}

\section{Introduction}
We study the problem of recovering a video of activity taking place outside our field of view by observing its indirect effect on shadows and shading in an observed region. This allows us to turn, for example, the pile of clutter in Figure~\ref{fig:setup} into a ``computational mirror'' with a low-resolution view into  non-visible parts of the room.

The physics of light transport tells us that the image observed on the clutter is related to the hidden image by a linear transformation. If we were to know this transformation, we could solve for the hidden video by matrix inversion (we demonstrate this baseline approach in Section~\ref{sec:inverselt}). Unfortunately, obtaining the transport matrix by measurement requires an expensive pre-calibration step and access to the scene setup. We instead tackle the hard problem of estimating \emph{both} the hidden video and the transport matrix simultaneously from a single input video of the visible scene. For this, we cast the problem as matrix factorization of the observed clutter video into a product of an unknown transport matrix and an unknown hidden video matrix. 

%Matrix factorization is known to be a very ill-posed problem. To get a factorization we can simply choose one of the factors at will, and recover a compatible companion factor by pseudoinversion. Consequently, most factorizations are almost completely arbitrary in terms of what the factor matrices represent. Optimization-based methods are prone to walking towards the nearest pair of matrices that produce the target matrix, which almost certainly is a meaningless local minimum.

%Matrix factorization is known to be a very ill-posed problem, as there are infinite possible factorizations for each different matrix. Optimization-based methods are prone to walking towards the nearest pair of matrices that multiply to the target matrix, which represents an almost certainly meaningless local minimum. 

Matrix factorization is known to be very ill-posed. Factorizations for any matrix are in principle easy to find: we can simply choose one of the factors at will (as a full-rank matrix) and recover a compatible %companion%removed word, companion matrix has a different specific meaning
factor by pseudoinversion. %\todo{should we discuss flip, rotation, oinear transforms of latent video?} 
Unfortunately, the vast majority of these factorizations are meaningless for a particular problem. %Optimization-based methods are prone to finding a nearby local minimum.
The general strategy for finding \emph{meaningful} factors is to impose problem-dependent priors or constraints --- for example, non-negativity or spatial smoothness. While successful in many applications, meaningful image priors can be hard to express computationally. In particular, we are not aware of any successful demonstrations of the inverse light transport factorization problem in the literature. We find that classical factorization approaches produce solutions that are scrambled beyond recognition. 

Our key insight is to build on the recently developed Deep Image Prior~\cite{ulyanov2018deep} to generate the factor matrices as outputs of a pair of convolutional neural networks trained in a one-off fashion. That is, we randomly initialize the neural net weights and inputs and then perform a gradient descent to find a set of weights such that their outputs' matrix product yields the known video. No training data or pre-training is involved in the process. Rather, the structure of convolutional neural networks, alternating convolutions and nonlinear transformations, induces a bias towards factor matrices that exhibit consistent image-like structure, resulting in recovered videos that closely match the hidden scene, although global ambiguities such as reflections and rotations remain. We found that this holds true of the video factor as well as the transport factor, in which columns represent the scene's response to an impulse in the hidden scene, and  exhibit image-like qualities.
\todo{Fredo asked to emphasize on the one-off training aspect of the method.}

% \footnote{
The source code, supplemental material, and a video demonstrating the results can be found at the project webpage at \url{compmirrors.csail.mit.edu}.

\section{Related Work}

%\todo{thanks for filling these in so far, looking good, though we'll need to be much more compact if we want to fit in the 8 page limit... at least remove the subsection titles?}

%This work lies at the intersection of a variety of different subfields of computer science, including matrix factorization, NLoS imaging, light transport measurement, and the Deep Image Prior. 

\textbf{Matrix Factorization.}
%The body of work in blind matrix factorization, sometimes also called blind source separation (BSS), is vast, and we cannot do it full justice given the space we have. However, we will briefly mention a few of the most important methods for matrix factorization, especially in the context of computer imaging. 
%In 1999, Lee and Seung used neural nets for non-negative imaging factorization as part of an effort to explain facial and object recognition in the human brain~\cite{lee1999learning}. More recently,~\cite{virtanen2007monaural} used unsupervised learning to do nonnegative matrix factorization for sound separation. Indeed, the use of neural nets and other learning techniques (such as MCMC or Bayes nets) for matrix factorization is a widespread phenomenon~\cite{hoyer2004non, sainath2013low, salakhutdinov2008bayesian, kompass2007generalized}.
Matrix factorization is a fundamental topic in computer science and mathematics. Indeed, many widely used matrix transformations and decompositions, such as the singular value decomposition, eigendecomposition, and LU decomposition, are instances of constrained matrix factorization. There has been extensive research in the field of blind or lightly constrained matrix factorization. The problem has applications in facial and object recognition~\cite{lee1999learning}, sound separation~\cite{virtanen2007monaural}, representation learning~\cite{trigeorgis2016deep}, and automatic recommendations~\cite{xue2017deep}. Neural nets have been used extensively in this field~\cite{hoyer2004non, sainath2013low, salakhutdinov2008bayesian, kompass2007generalized}, and are often for matrix completion with low-rank assumption~\cite{trigeorgis2016deep, xue2017deep}.

% Like our method, these latter methods learn from a single input matrix, but unlike ours, they do not assume image-like structure within the matrix.
%\todo{Discuss low rank. discuss how prior neural methods relate to ours.}

Blind deconvolution~\cite{freeman2009understanding, krishnan2011blind, fergus2006removing, levin2011efficient,schulz1993multiframe, jefferies1993restoration} is closely related to our work but involves a more restricted class of matrices. This greatly reduces the number of unknowns (a kernel rather than a full matrix) and makes the problem less ill-posed, although still quite challenging.

Koenderink et al.~\cite{Koenderink1997} analyze a class of problems where one seeks to simultaneously estimate some property, and calibrate the linear relationship between that property and the available observations. Our blind inverse light transport problem falls into this framework. \todo{removed the tail of this sentence, I think it misses the point (see commented part). could maybe think of something better to say.} %, as we also seek to find a pair of matrices with a known matrix product that have image-like structure.
%\todo{Say more how it relates: known structure of matrix, simpler.}

%Blind deconvolution methods are of interest to us because of their frequent application in imaging systems, including NLoS ones, and because they represent a special case of the matrix factorization problem: one where all relevant matrices have Toeplitz-like structure.

%\todo{someone could literally google for ``neural matrix factorization'' or 'deep matrix factorization' or so on, and see what those few papers are about. they seem to address non-image stuff, but good to mention. perhaps they even use similar ideas as we do.}

\textbf{Deep Image Prior.} In 2018, Ulyanov et al.~\cite{ulyanov2018deep} published their work on the Deep Image Prior---the remarkable discovery that due to their structure, convolutional neural nets inherently impose a natural-image-like prior on the outputs they generate, even when they are initialized with random weights and without any pretraining. 
%This stunning result is the direct inspiration for the method we present in our paper.
Since the publication of~\cite{ulyanov2018deep}, there have been several other papers that make use of the Deep Image Prior for a variety of applications, including compressed sensing~\cite{van2018compressed}, image decomposition~\cite{gandelsman2018double}, denoising~\cite{cheng2019bayesian}, and image compression~\cite{heckel2018deep}. In concurrent work, the Deep Image Prior and related ideas have also been applied to blind deconvolution \cite{Asim2018,Ren2019}.

%who find the Deep Image Prior to be equivalent to a stationary Gaussian process and use the method for denoising and inpainting. Finally, Heckel and Hand~\cite{heckel2018deep} show that untrained neural networks can be used to more concisely represent or compress natural images. 

%\MA{missing double-dip paper, it's a direct follow-up}

%\MA{couple of papers: DEEP DECODER:CONCISE IMAGE REPRESENTA-TIONS FROM UNTRAINED NON-CONVOLUTIONAL NETWORKS 
%.
%Also: A Bayesian Perspective on the Deep Image Prior }

\textbf{Light Transport Measurement.}
There has been extensive past work on measuring and approximating light transport matrices using a variety of techniques, including compressed sensing~\cite{peers2009compressive}, kernel Nystr{\"o}m methods~\cite{wang2009kernel}, and Monte Carlo methods~\cite{sharma2003role}. \cite{sen2005dual} and~\cite{garg2006symmetric} study the recovery of an image's reflectance field, which is the light transport matrix between the incident and exitant light fields.

%To our knowledge, no prior work tackles the blind problem of estimating light transport from a single uncontrolled video. 

%\todo{cite dual photography and follow up. Symmetric Photography: Exploiting Data-sparseness in Reflectance Fields. Mention reflectance fields.}
%To our knowledge, no prior work has used the Deep Image Prior to approximate light transport matrices.

%[\textbf{Is this section necessary? I'm not sure ``light transport measurement'' is a real/coherent subfield... Happy to be overruled of course...} -Adam]

%\MA{There is a bunch of literature that explicitly works with measuring or analyzing these matrices/operators, e.g. Compressive Light Transport Sensing ; Kernel Nystrom for Light Transport ; maybe A Theory of Locally Low Dimensional Light Transport ; etc, so it might be worthwhile to discuss them briefly}

\textbf{Non-Line-of-Sight (NLoS) Imaging.}
%Past work in NLoS imaging can be split into one of two categories. The first is \textit{active} methods, which introduce light into the scene and use known properties about that light (such as phase or time of arrival) to infer information about the hidden scene. The second is \textit{passive} methods, which make use of visible reflections from ambient light in the scene to infer information about the hidden scene. 
%Although our experimental results are derived from projected images on a wall, we believe our methods bear a stronger resemblance to previously-studied passive NLoS systems than they do to active systems. In principle, our methods could be applied to perform imaging on scenes that are ambiently rather than actively lit, and do not require time-of-flight information or laser illumination to work. 
%\PS{Remove the subsubsection to be consistent with the format. }
%\subsubsection{Active methods}
Past work in \emph{active} NLoS imaging focuses primarily on active techniques using time-of-flight information to resolve scenes~\cite{xu2016photon, shrestha2016computational, heide2014diffuse}. Time-of-flight information allows the recovery of a wealth of information about hidden scenes, including number of people~\cite{xia2011human}, object tracking~\cite{klein2016tracking}, and general 3D structure~\cite{gariepy2016detection, pandharkar2011estimating, velten2012recovering}. In contrast, past work in \emph{passive} NLoS imaging has focused primarily on occluder-based imaging methods. These imaging methods can simply treat objects in the surrounding environment as pinspecks or pinholes to reconstruct hidden scenes, as in~\cite{Torralba2014} or~\cite{saunders2019computational}. Others have used corners to get 1D reconstructions of moving scenes~\cite{bouman2017turning}, or used sophisticated models of occluders to infer light fields~\cite{baradad2018inferring}.

\section{Inverse Light Transport \label{sec:inverselt}}

We preface the development of our factorization method by introducing the inverse light transport problem, and presenting numerical real-world experiments with a classical matrix inversion solution when the transport matrix is known. In later sections we study the case of \emph{unknown} transport matrices.

\subsection{Problem Formulation}

\begin{figure*}[bt]
\centering
\includegraphics[width=\linewidth]{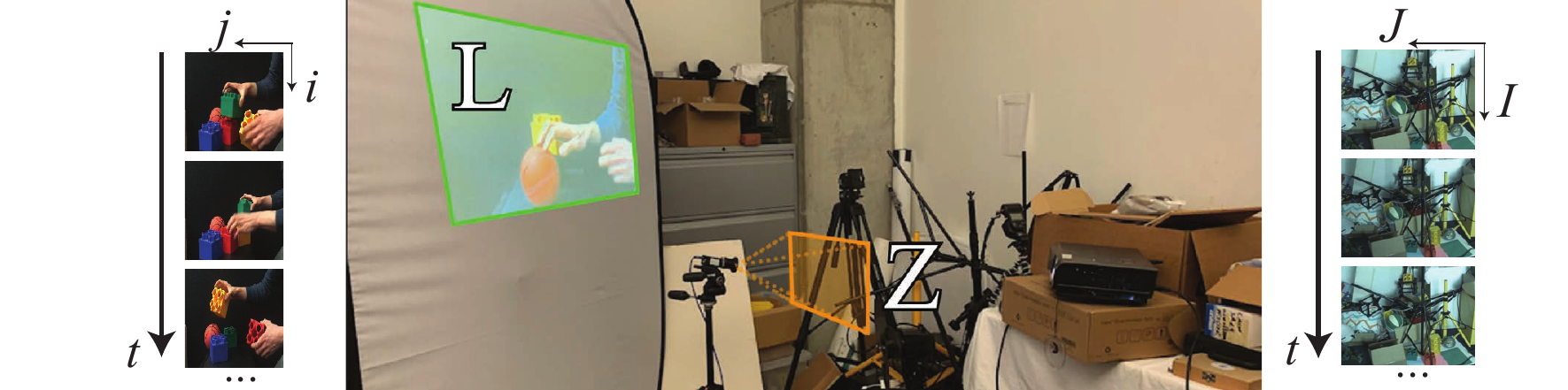}
\caption{
A typical experimental setup used throughout this paper. A camera views a pile of clutter, while a hidden video $L$ is being projected outside the direct view $Z$ of the camera. We wish to recover the hidden video from the shadows and shading observed on the clutter. The room lights in this photograph were turned on for the purpose of visualization only. During regular capture, we try to minimize any sources of ambient light.
%The camera is recording a static scene, in this case consisting of a clutter of objects. Hidden from direct view of the camera, a video is being projected onto the wall. The motion in the hidden video affects the lighting conditions in the observed region, causing subtle changes in shadows and shading that are indicative of the activity in the video. We address the problem of recovering the hidden video $L$ from the recorded observations $Z$. The dimensions of these tensors are illustrated in the figure. 
We encourage the reader to view the supplemental video to see the data and our results in motion.
%The recorded video frames constitute a 3D tensor $Z \in \real^{I \times J \times t}$, where $I$ and $J$ are the image dimensions, and $t$ is the frame number. 
}
\label{fig:setup}
\end{figure*}

%\todo{this came out way too long again, and probably way too much philosophy considering what is needed to understand the method later. I wonder if the whole thing could be just explained as more of a special case of how i*j projector pixels each have a response on the observation and that it's a linear thing. But on the other hand don't want to make it sound overly constrained to just this unrealistic setup...}

The problem addressed in this paper is illustrated in Figure \ref{fig:setup}. We observe a video $Z$ of, for example, a pile of clutter, while a hidden video $L$ plays on a projector behind the camera. Our aim is to recover $L$ by observing the subtle changes in illumination it causes in the observed part of the scene. Such cues include e.g. shading variations, and the clutter casting moving shadows in ways that depend on the distribution of incoming light (i.e. the latent image of the surroundings). 
%We stress that our method is \emph{not} based on e.g. analyzing the silhouettes of direct shadows cast by external objects moving outside the frame; rather, the clutter casts shadows on itself and the back wall, providing indirect cues about the light (i.e. the image) from its surroundings. 
The problem statement discussed here holds in more general situations than just the projector-clutter setup, but we focus on it to simplify the exposition and experimentation.

%We also completely sidestep any explicit estimation of geometry or reflectance properties of the scene by focusing on the algebraic aspects of the problem.

%The problem addressed in this paper is illustrated in Figure \ref{fig:setup}. We observe a video $Z$ of, for example, a pile of clutter. The camera and the observed objects are static, but outside the field of view of the camera, a hidden video $L$ is being projected onto a wall. The change in surrounding illumination causes subtle motion in shadows and shading of $Z$, providing cues about the activity taking place outside the view. Our goal is to recover the hidden video by inverting the linear light transport transformation that relates pixels in $Z$ to pixels in $L$. Note that this problem statement and analysis extends beyond just the projector setup, but for the purposes of simplifying the exposition and controlling our experiments, we describe the image formation explicitly assuming this scenario.
The key property of light transport we use is that it is \emph{linear}~\cite{kajiya1986rendering, pharr2016physically}: if we light up two pixels on the hidden projector image in turn and sum the respective observed images of the clutter scene, the resulting image is the same as if we had taken a single photograph with both pixels lit at once. More generally, for every pixel in the hidden projector image, there is a corresponding response image in the observed scene. When an image is displayed on the projector, the observed image is a weighted sum of these responses, with weights given by the intensities of the pixels in the projector image. In other words, the image formation is the matrix product
\begin{equation}
\label{eq:transport}
Z = TL
\end{equation}
%where $Z \in \real^{IJ}$ is the observed image of resolution $I*J$,  and $L \in \real^{ij}$ is the hidden image of resolution $i*j$, and the light transport matrix $T \in \real^{IJ \times ij}$ contains all of the response images in its columns.\footnote{Throughout this paper, we use notation such as $T \in \real^{IJ \times ij}$ to imply that $T$ is, in principle, a 4-dimensional tensor with dimensions $I$, $J$, $i$ and $j$, and that we have packed it into a 2-dimensional tensor (matrix) by stacking the $I$ and $J$ dimensions together into the columns, and $i$ and $j$ in the rows. This corresponds to the \texttt{view()} command in PyTorch. Thus, when we refer to e.g. columns of this matrix as images, we are really referring to the unpacked $I \times J$ array corresponding to that column.}
where $Z \in \real^{IJ \times t}$ is the observed image of resolution $I*J$ at $t$ time instances, and $L \in \real^{ij \times t}$ is the hidden video of resolution $i*j$ (of same length $t$), and the light transport matrix $T \in \real^{IJ \times ij}$ contains all of the response images in its columns.\footnote{Throughout this paper, we use notation such as $T \in \real^{IJ \times ij}$ to imply that $T$ is, in principle, a 4-dimensional tensor with dimensions $I$, $J$, $i$ and $j$, and that we have packed it into a 2-dimensional tensor (matrix) by stacking the $I$ and $J$ dimensions together into the columns, and $i$ and $j$ in the rows. Thus, when we refer to e.g. columns of this matrix as images, we are really referring to the unpacked $I \times J$ array corresponding to that column.}
%As we are working with video, we simply expand $Z$ and $L$ with a time dimension, i.e. $Z \in \real^{IJ \times t}$ and $L \in \real^{ij \times t}$, and apply the same equation. 
$T$ has no dependence on time, because the camera, scene geometry, and reflectance are static.

This equation is the model we work with in the remainder of the paper. %this section and Section~\ref{sec:DIPbasedMF}. %Its components are summarized and illustrated in Figure TODO. 
In the subsequent sections, we make heavy use of the fact that all of these matrices exhibit image-like spatial (and temporal) coherence across their dimensions. %\todo{Fredo: Maybe change continuity to coherence}
% A useful intuition for understanding the matrix $T$ is that by viewing each of its columns in turn, one would see images of the shadows cast by the clutter moving in straight lines, as though there were a moving lamp outside the frame.
A useful intuition for understanding the matrix $T$ is that by viewing each of its columns in turn, one would see the scene as if illuminated by just a single pixel of the hidden image.
%if one were to view each of its column images in turn, the ``video'' would look as though a small point light source was doing a raster scan over a rectangular region, causing sharp shadows in the observed image to move in straight lines. Conversely, the row images reveal, for any pixel in the observed image, which pixels in the projector are visible to that point. In the mixed dimensions one would see elongated slanted lines corresponding to the motion of shadows. Please see the supplemental video for an illustration of these components.

%More generally, the hidden ``light sources'' described by $L$ need not be actual light emitters. They can just as well represent the spherical illumination field the scene receives from reflected light in its environment (in other words, a ``fisheye image'' of the scene's surroundings). This relation is, however, only approximately linear when the distance between the observed and hidden scene is finite.

\subsection{Inversion with Known Light Transport Matrix}

We first describe a baseline method for inferring the hidden video from the observed one in the non-blind case. In addition to the observed video $Z$, we assume that we have previously measured the light transport $T$ by lighting up the projector pixels individually in sequence and recording the response in the observed scene. We discretize the projector into $32 \times 32$ square pixels, corresponding to $i = j = 32$. As we now know two of the three matrices in Equation \ref{eq:transport}, we can recover $L$ as a solution to the least-squares problem $\mathrm{argmin}_L ||TL - Z||_2^2$. We augment this with a penalty on spatial gradients. In practice, we measure $T$ in a mixed DCT and PCA basis to improve the signal-to-noise ratio, and invert the basis transformation post-solve. We also subtract a black frame with all projector pixels off from all measured images to eliminate the effect of ambient light.

\begin{figure*}[bt]
\centering
\includegraphics[width=\linewidth]{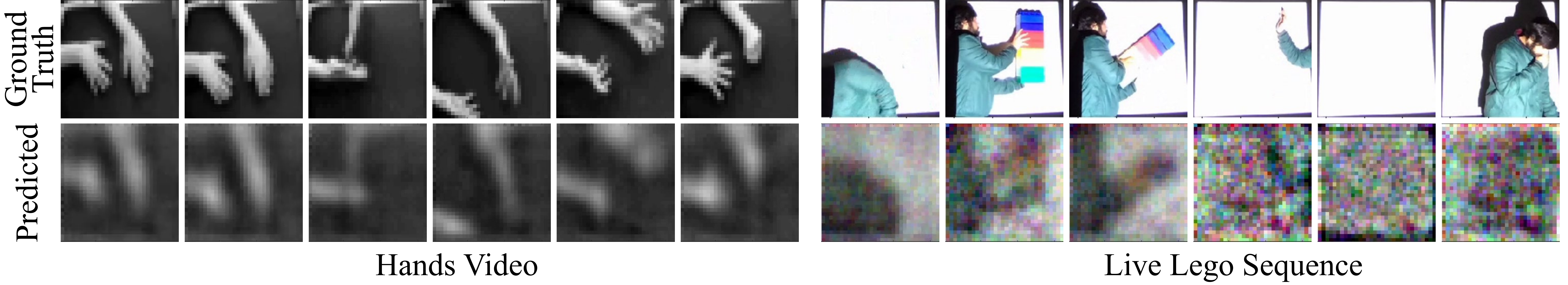}
\caption{
Reconstructions with known light transport matrix.
%\MA{the top row is a bit useless in paper form, because the frames look almost identical, maybe refer to Fig 1 and say that that's what they look like}
%\AY{seems to me that's a good thing; it implies that the method must be sophisticated in order to get reconstructions from such subtle differences}
%\PS{Will eat up space and we are going to show stuff in the video to make that point anyways.} \MA{I do see adam's argument, but still it's kind of problematic because the tininess and stillness of the images hides the action... maybe if it were a zoom into some region, but I don't know}
}
\label{fig:known_lt_results}
\end{figure*}

Figure~\ref{fig:known_lt_results} shows the result of this inversion for two datasets, one where a video was projected on the wall, and the other in a live-action scenario. The recovered video matches the ground truth, although unsurprisingly it is less sharp. This non-blind solution provides a sense of  upper bound for the much harder blind case.

The amount of detail recoverable depends on the content of the observed scene and the corresponding transport matrix \cite{Mahajan2007}. In this and subsequent experiments, we consider the case where %\todo{Maybe assume --> consider} %a sufficient amount of geometric complexity to generate high-frequency 
%features like shadows cast by the observed geometry onto itself, 
 the objects in the observed scene have some mutual occlusion, so that they cast shadows onto one another. This improves the conditioning of the light transport matrix. In contrast, for example, a plain diffuse wall with no occluders would act as a strong low-pass filter and eliminate the high-frequency details in the hidden video. However, we stress that we do not rely on explicit knowledge of the geometry, nor on \emph{direct} shadows being cast by the hidden scene onto the observed scene.

%except for high frequency detail that is lost to the light transport and measurement inaccuracies.

%Figure~\ref{fig:known_lt_results} shows a challenging live action scenario where some of our assumptions are violated due to mismatch in implicitly assumed depth and misalignment with the basis. Here, the projector was simply showing a white frame, and a person performed motions in the illuminated region during the capture. We find that the inversion is still able to recover a surprising amount of detail in this video. %The color channels are solved for separately. \PS{Maybe the color detail is not required to be said explicitly.}

%, despite the person no longer being quite properly aligned with the measured basis, due to mismatch with implicitly assumed depth. For viewing purposes, the video has been centered to unit brightness and standard deviation on a per-frame basis; this seems to effectively suppress the most serious flickering artifacts caused by these violations to the model. We obtained the colored video by simply solving for each channel independently.

%The remainder of this paper is concerned with the significantly more difficult case where calibration is not available, and \emph{both} $L$ and $T$ must be simultaneously estimated from $Z$ via matrix factorization.
\section{Deep Image Prior based Matrix Factorization \label{sec:DIPbasedMF}}
%In this section, we describe a Deep Image Prior based method which takes in product of two 2D matrices and factors them into the possible two factor matrices.
Our goal is to recover the latent factors when we \emph{do not} know the  light transport matrix. In this section, we describe a novel matrix factorization method that uses the Deep Image Prior \cite{ulyanov2018deep} to encourage natural-image-like structure in the factor matrices. We first describe numerical experiments with one-dimensional toy versions of the light transport problem, as well as general image-like matrices. We also demonstrate the failure of classical methods to solve this problem. Applications to real light transport configurations will be described in the next section. 

\subsection{Problem Formulation}
In many inference problems, it is known that the observed quantities are formed as a product of latent matrices, and the task is to recover these factors. Matrix factorization techniques seek to decompose a matrix $Z$ into a product $Z \approx TL$ (using our naming conventions from Section~\ref{sec:inverselt}), either exactly or approximately. The key difficulty is that a very large space of different factorizations satisfy any given matrix. 

%\PS{Minor but replace the word massive maybe to something more formal.}
%, but most of them are meaningless for any given application. 
%Indeed, we can trivially obtain a factorization by choosing an arbitrary full-rank matrix (of compatible dimensions) for one of the factors, and then recovering the other factor by multiplying $Z$ with its pseudoinverse. Therefore merely finding \emph{a} factorization is meaningless for most applications.  
%In particular, the singular value decomposition (SVD) can always be used to decompose $Z = U\Sigma V^T$, and to set the factors as $T = U \sqrt{\Sigma}$ and $L = \sqrt{\Sigma}V^T$. Furthermore, given a pair of valid factors $T$ and $L$, the factors $TQ^{-1}$ and $QL$ also represent a valid factorization, for any suitably shaped full-rank choice of $Q$, as $(TQ^{-1})(QL) = TL = Z$. [cite koenderink]. Note that, conversely, this also characterizes the space of valid factorizations -- we will make use of this fact in Section TODO.
The most common solution is to impose additional priors on the factors, for example, non-negativity~\cite{lee1999learning, virtanen2007monaural} and spatial continuity~\cite{levin2011efficient}. They are tailored on a per-problem basis, and often capture the desired properties (e.g. likeness to a natural image) only in a loose sense. Much of the work in nonnegative matrix factorization assumes that the factor matrices are low-rank rather than image-like.

%To distinguish between the equally well fitting factorizations, a typical approach is to specify \emph{priors} on the factors. Our interest is image-related problems, where it is typical to specify that images should be non-negative, leading to a wide range of non-negative matrix factorization (NNMF) techniques [CITE]. To enforce continuity across dimensions that have a spatial or temporal character, smoothness priors such as Total Variation (TV) are commonly used [CITE]. These priors are typically tailored for the problem at hand, and achieve varying levels of success. %Generally speaking, assuming that a the matrix $Z$ is simply a product of two matrices that upon visual inspection appear as ``natural images'', these classical priors tend to be too weak to reliably recover meaningful structure in the factors. We will explore this in Section TODO.

%\todo{I like this viewpoint but maybe need to condense to half a sentence...}
%\PS{Should this be discussed in the discussion section under consistency under repeated trials? }
The combination of being severely ill-posed and non-convex makes matrix factorization highly sensitive to not only the initial guess, but also the \emph{dynamics} of the optimization. While this makes analysis hard, it also presents an opportunity: by shaping the loss landscape via suitable parameterization, we can guide the steps towards better local optima. This is a key motivation behind our method.

\subsection{Method}
\label{sec:DIPmethod}

\begin{figure*}[bt]
\centering
\includegraphics[width=\linewidth]{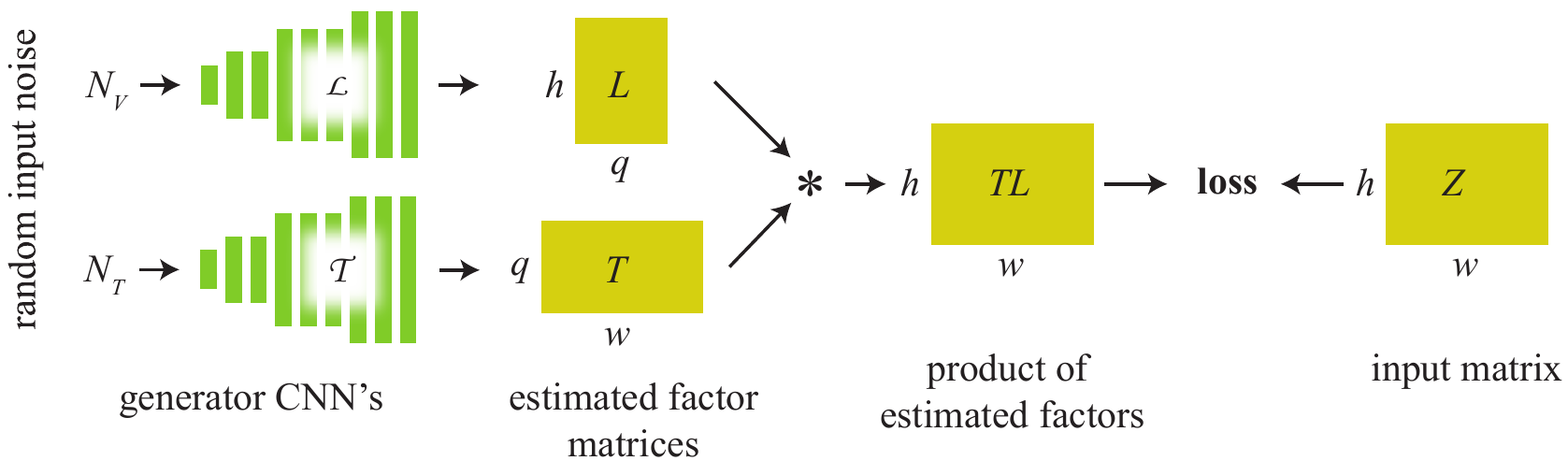}
\caption{
High level overview of our matrix factorization approach. The CNNs are initialized randomly and ``overfitted'' to map two vectors of noise onto two matrices $T$ and $L$, with the goal of making their product match the input matrix $Z$. In contrast to optimizing directly for the entries of $T$ and $L$, this procedure regularizes the factorization to prefer image-like structure in these matrices.
}
\label{fig:architecturesimple}
\end{figure*}

We are inspired by the Deep Image Prior~\cite{ulyanov2018deep} and Double-DIP~\cite{gandelsman2018double}, where an image or a pair of images is parameterized via convolutional neural networks that are optimized in a one-off manner for each test instance. We propose to use a pair of one-off trained CNNs to generate the two factor matrices in our problem (Figure~\ref{fig:architecturesimple}). We start with two randomly initialized CNNs, each one outputting a respective matrix $L$ and $T$. Similarly to~\cite{ulyanov2018deep}, these CNNs are not trained from pairs of input/output labeled data, but are trained only once and specifically to the one target matrix. The optimization adjusts the weights of these networks with the objective of making the product of their output matrices identical to the target matrix being factorized. 
The key idea is that the composition of convolutions and pointwise nonlinearities has an inductive bias towards generating image-like structure, and therefore is more likely to result in factors that have the appearance of natural images. The general formulation of our method is the minimization problem
\begin{equation}
\label{eq:general_formulation}
    \mathrm{argmin}_{\theta, \phi} d(\mathcal{T}(N_T; \theta) \mathcal{L}(N_L; \phi), Z)
\end{equation}
where $Z \in \real^{h \times w}$ is the matrix we are looking to factorize, and $\mathcal{T}: \real^{n_T} \mapsto \real^{h \times q}$ and $\mathcal{L}: \real^{n_L} \mapsto \real^{q \times w}$ are functions implemented by convolutional neural networks, parametrized by weights $\theta$ and $\phi$, respectively. These are the optimization variables. $q$ is a chosen inner dimension of the factors (by default the full rank choice $q = \mathrm{min}(w,h)$). $d: \real^{w \times h} \times \real^{w \times h} \mapsto \real$ is any problem-specific loss function, e.g. a pointwise difference between matrices. The inputs $N_T \in \real^{n_T}$ and $N_L \in \real^{n_L}$ to the networks are typically fixed vectors of random noise. Optionally the values of these vectors may be set as learnable parameters. They can also subsume other problem-specific inputs to the network, e.g. when one has access to auxiliary images that may guide the network in performing its task. The exact design of the networks and the loss function is problem-specific.
%, and different architectural choices can be used to steer the behavior of the networks, and optionally to target the image-like structure to a desired subset of dimensions. In the following subsection we present one instance of this scheme. In Section TODO, we use a more complex architecture tailored for the 3D and 4D tensors involved in the full light transport problem.

\subsection{Experiments and Results}

%\paragraph{1D Light Transport Toy Model} 
%We test the CNN-based framework on a set of factorization tasks, where we created 1D toy data that has a similar character with the light transport factorization problem. We formed the ground truth by matrix multiplication and sought to recover the factors using the method discussed above.
We test the CNN-based factorization approach % framework %\todo{Fredo doesn't like the word framework}
on synthetically generated tasks, where the input is a product of a pair of known ground truth matrices. We use both toy data that simulates the characteristics of light transport and video matrices, as well as general natural images.
%: a matrix $T_0$ with slanted edges, similar to mixed-dimensional slices of real transport matrices, and a 
%: a  matrix $T_0 \in \real^{I \times i}$ consisting of horizontal and diagonal edges (superficially reminiscent of a light field), and a matrix $L_0 \in \real^{i \times t}$ representing a 1D video of dots moving over the time dimension. Observe that both of these matrices have the appearance of being a 2-dimensional image. From these, we form the method input $Z = T_0 L_0$. The goal is to factorize $Z$ into matrices $T$ and $L$ that appear similar to the ground truth $T_0$ and $L_0$.

We design the generator neural networks $\mathcal{T}$ and $\mathcal{L}$ as identically structured sequences of convolutions, nonlinearities and upsampling layers, %illustrated in Figure~\ref{fig:architecturefull}.  
detailed in the supplemental appendix. \todo{restored old reference to supplemental, there was an incorrect reference to fig 5 here. maybe just remove though, see repeated reference below?}
To ensure non-negativity, the final layer activations are exponential functions. Inspired by~\cite{liu2018intriguing}, we found it useful to inject the pixel coordinates as auxiliary feature channels on every layer. Our loss function is $d(x,y) = ||\nabla (x-y)||_1 + w ||x-y||_1$ where $\nabla$ is the finite difference operator along %$i$ and $j$
the spatial dimensions, and $w$ is a small weight; the heavy emphasis on local image gradients appears to aid the optimization. 
We use Adam~\cite{kingma2014adam} as the optimization algorithm. The details can be found in the supplemental appendix.

Figure~\ref{fig:factorization1d} shows a pair of factorization results, demonstrating that our method is able to extract images similar to the original factors. We are not aware of similar results in the literature; as a baseline we attempt these factorizations with the DIP disabled, and with standard non-negative matrix factorization. These methods fail to produce meaningful results. %Despite reasonable efforts, we were unable to obtain meaningful results with these methods. 
% In the supplemental material %\todo{verify that it will be there in time} 
% we study the robustness of the method to injected noise, and find a graceful drop in performance.

%We also test the method on inputs formed as products of pairs of photographs interpreted as matrices, and found success in particular with images exhibiting high-contrast geometric features. While this task doesn't directly model any meaningful problem, it is demonstrative of the power of our method. %, and suggests that our DIP-based factorization has potential for general use beyond the light transport problem. 

%\paragraph{Robustness} Many inverse image problems, such as (blind) deconvolution, are known to be very sensitive to noise. In Figure TODO, we study the robustness of our approach to noise that is added on the input matrix $Z$. In general, we find graceful drop of performance as a function of the noise level. TODO
\begin{figure*}[bt]
\centering
\includegraphics[width=\linewidth]{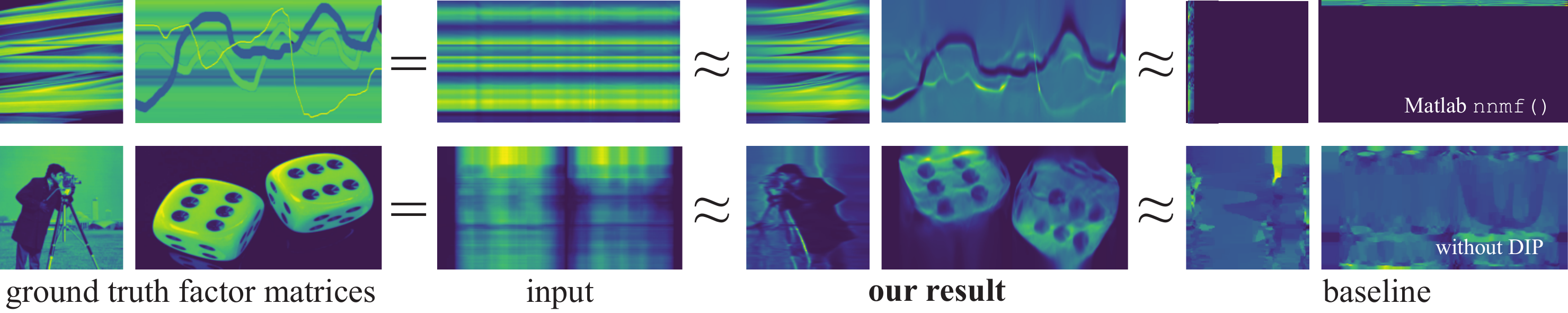}
\caption{
Matrix factorization results. The input to the method is a product of the two leftmost matrices. Our method finds visually readable factors, and e.g. recovers all three of the faint curves on the first example. On the right, we show two different baselines: one computed with Matlab's non-negative matrix factorization (in alternating least squares mode), and one using our code but optimizing directly on matrix entries instead of using the CNN, with an $L_1$ smoothness prior.
}
\label{fig:factorization1d}
\end{figure*}

\subsection{Distortions and Failure Modes}
\label{sec:distortions}
The factor matrices are often warped or flipped. % \todo{maybe we should discuss this ambiguity much earlier.} 
This stems from ambiguities in the factorization task, as the factor matrices can express mutually cancelling distortions. However, the DIP tends to strongly discourage distortions that break spatial continuity and scramble the images.

More specifically, the space of ambiguities and possible distortions can be characterized as follows \cite{Koenderink1997}. Let $T_0$ and $L_0$ be the true underlying factors, the observed video 
thus being $Z = T_0 L_0$. All valid factorizations are then of the form $T = T_0 A^\dagger$ and $L = A L_0$,  
where $A$ is an arbitrary full-rank matrix, and $A^\dagger$ is its inverse. This can be seen by 
substituting $TL = (T_0 A^\dagger)(AL_0) = T_0 (A^\dagger A)L_0 = T_0 L_0 = Z$. 

The result of any factorization implicitly corresponds to some choice of $A$ (and $A^\dagger$). 
In simple and relatively harmless cases (in that they do not destroy the legibility of the images), the matrix $A$ can represent e.g. a permutation that flips the image, 
whence $A^\dagger$ is a flip that restores the original orientation. %: this case is illustrated 
%in Figure 4’s conversely flipped matrices. 
They can also represent reciprocal intensity modulations, meaning that there is a fundamental ambiguity about the intensity of the factors. However, for classical factorization methods, the matrices tend to consist of unstructured 
``noise'' that scrambles the image-like structure in $T_0$ and $L_0$ beyond recognition. Our finding is that the use of DIP discourages such factorizations.

\section{Blind Light Transport Factorization \label{sec:blind_dip_lt}}
We now combine the ideas from the previous two sections and present a method for tackling the inverse light transport problem blindly, when we have no access to a measured light transport matrix. We show results on both synthetic and real data, and study the behavior of the method experimentally.

\subsection{Method}
\begin{figure*}[bt]
\centering
\includegraphics[width=\linewidth]{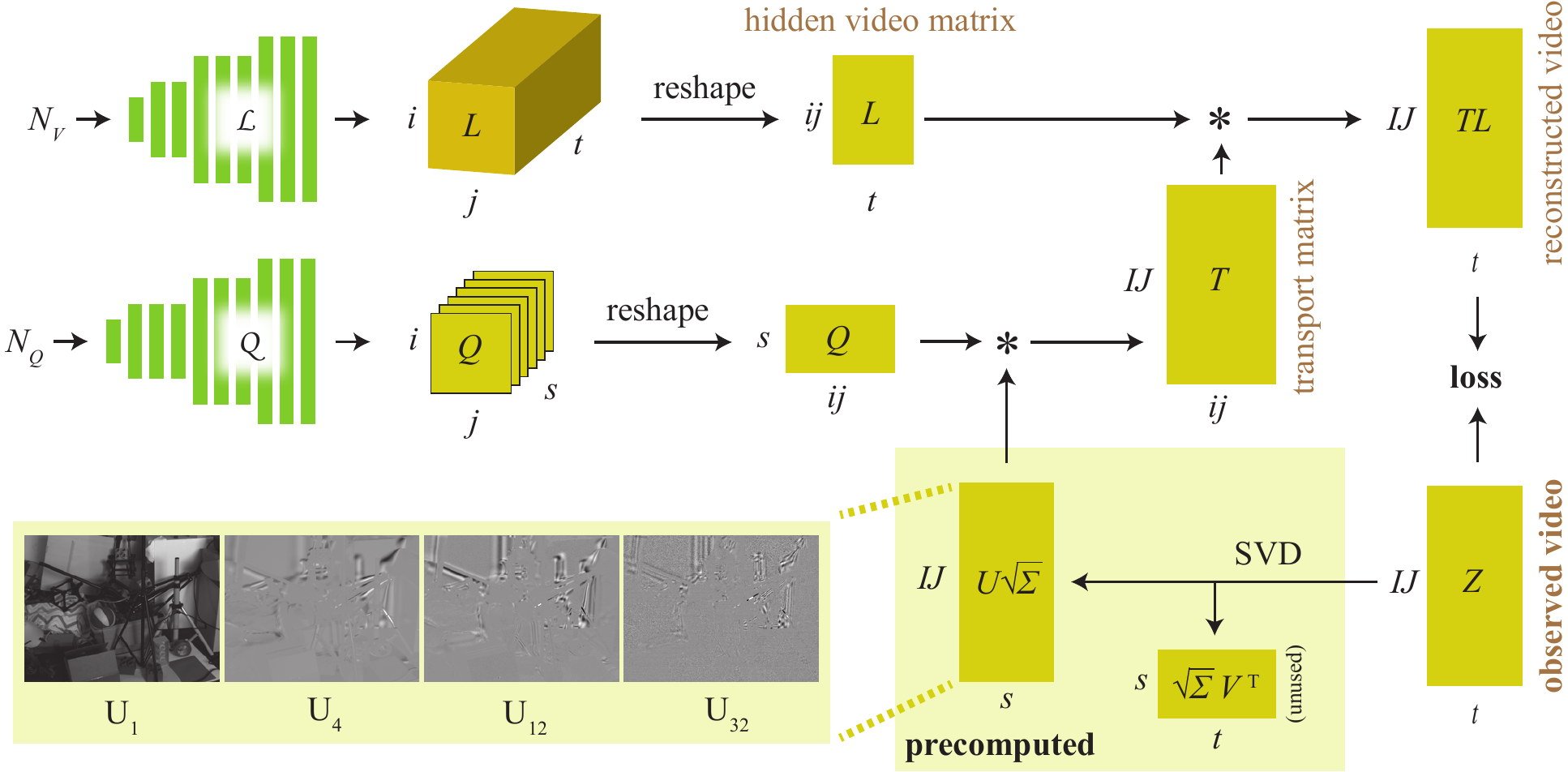}
\caption{
An overview of the architecture and data flow of our blind inverse light transport method. Also shown (bottom left) are examples of the left singular vectors stored in $U$. $\mathcal{L}$ and $\mathcal{Q}$ are convolutional neural networks, and the remainder of the blocks are either multidimensional tensors or matrices, with dimensions shown at the edges. The matrices in the shaded region are computed once during initialization. The input $Z$ to the method is shown in the lower right corner.
%\PS{Maybe add a table based network architecture. }
}
\label{fig:architecturefull}
\end{figure*}

\paragraph{Setup} Continuing from Section~\ref{sec:inverselt}, our goal is to factor the observed video $Z \in \real^{IJ \times t}$ of $I * J$ pixels and $t$ frames, into a product of two matrices: the light transport $T \in \real^{IJ \times ij}$, and the hidden video $L \in \real^{ij \times t}$. The hidden video is of resolution $i*j$, with $i = j = 16$. %for reasons of computational efficiency, and the limited expected resolving power. 
Most of our input videos are of size $I=96$ (height), $J=128$ (width), and $t$ ranging from roughly 500 to 1500 frames.

Following our approach in Section~\ref{sec:DIPbasedMF}, the task calls for designing two convolutional neural networks that generate the respective matrices. Note that $T$ can be viewed as a 4-dimensional $I \times J \times i \times j$ tensor, and likewise $L$ can be seen as a 3-dimensional $i \times j \times t$ tensor. We design the CNNs to generate the tensors in these shapes, and in a subsequent network operation reshape the results into the stacked matrix representation, so as to evaluate the matrix product. The dimensionality of the convolutional filters determines which dimensions in the result are bound together with image structure.
% filters can freely operate across chosen dimensions to induce the desired image structure. 
In the following, we describe the networks generating the factors. An overview of our architecture is shown in Figure \ref{fig:architecturefull}.

\paragraph{Hidden Video Generator Network} The hidden video tensor $L$ should exhibit image-like structure along all of its three dimensions. Therefore a natural model is to use 3D convolutional kernels in the network $\mathcal{L}$ that generates it. Aside from its dimensionality, the network follows a similar sequential up-scaling design as that discussed in Section~\ref{sec:DIPbasedMF}. It is illustrated in Figure~\ref{fig:architecturefull} and detailed in the supplemental appendix.

\paragraph{Light Transport Generator Network} The light transport tensor $T$, likewise, exhibits image structure between all its dimensions, which in principle would call for use of 4D convolutions. Unfortunately these are very slow to evaluate, and unimplemented in most CNN frameworks. We also initially experimented with alternating between 2D convolutions along $I,J$ dimensions and $i,j$ dimensions, and otherwise following the same sequential up-scaling design. While we reached some success with this design, we found a markedly different architecture to work better.

The idea is to express the slices of $T$ as linear combinations of basis images obtained from the singular value decomposition (SVD) of the input video. This is both computationally efficient and guides the optimization by constraining the iterates and the solution to lie in the subspace of valid factorizations. %For computational efficiency and guiding the optimization towards the right path, we seek to prune the search space by constraining the solved basis images in $T$ to be in the subspace of images that characterizes the possible factorizations of $Z$. %\PS{Maybe have a bridging sentence about "subspace of images from SVD being a good idea".} 
Intuitively, the basis expresses a frequency-like decomposition of shadow motions and other effects in the video, as shown in Figure \ref{fig:architecturefull}. 
%and linear combinations of the most important singular vectors should be able to express the actual physical basis images we seek for $T$. We will see below that this is indeed the case.

We begin by precomputing the truncated singular value decomposition $U \Sigma V^T$ of the input video $Z$ (with the highest $s = 32$ singular values), and aim to express the columns of $T$ as linear combinations of the left singular vectors $U \in \real^{IJ \times s}$. The individual singular vectors have the dimensions $I \times J$ of the input video. These vectors form an appropriate basis for constructing the physical impulse response images in $T$, as the the column space of $Z$ coincides with that of $T$ due to them being related by right-multiplication.
%, as $Z$ is (both computationally and physically) formed by \emph{right}-multiplying $T$ with another matrix.
\footnote{Strictly speaking, some dimensions of the true $T$ may be lost in the numerical null space of $Z$ (or to the truncated singular vectors) if the light transport ``blurs'' the image sufficiently, making it impossible to exactly reproduce the $T$ from $U$. In practice we find that at the resolutions we realistically target, this does not prevent us from obtaining meaningful factorizations.}

We denote the linear combination by a matrix $Q \in \real^{s \times ij}$. The task boils down to finding $Q$ such that $(UQ) L \approx Z$. Here $L$ comes from the DIP-CNN described earlier. While one could optimize for the entries of $Q$ directly, we again found that generating $Q$ using a CNN produced significantly improved results. For this purpose, we use a CNN that performs 2D convolutions in the $ij$-dimension, but \emph{not} across $s$, as only the former dimension is image-valued. In summary, the full minimization problem becomes a variant of Eq.~\ref{eq:general_formulation}:
\begin{equation}
\label{eq:full_formulation}
    \mathrm{argmin}_{\theta, \phi} d(U \sqrt{\Sigma} \mathcal{Q}(N_Q; \theta) \mathcal{L}(N_L; \phi), Z)
\end{equation}
where $\mathcal{Q}$ implements the said CNN. The somewhat inconsequential additional scaling term $\sqrt{\Sigma}$ originates from our choice to distribute the singular value magnitudes equally between  the left and right singular vectors.

%, and we are constructing $T = (U\sqrt{\Sigma}Q)$ by optimizing for the much smaller matrix $Q$.

%of this linear combination are expressed by a matrix $Q \in \real^{s \times ij}$ that is generated by a DIP-based CNN that imposes image structure along the $ij$-dimension, but \emph{not} along $s$.

%; this captures most action in the video, as light transport tends to be low dimensional, similar to a blur

%Recall (Section TODO) that the space of possible factorizations for $Z$ is fully \todo{verify that it was fully} spanned by matrices $T_Q = U \sqrt{\Sigma}Q^{-1}$ and $L_Q = Q\sqrt{\Sigma}V^T$, where $Q$ is an arbitrary full rank square matrix of compatible shape, and $U \Sigma V^T$ is the singular value decomposition of $Z$. Let us hereafter the truncated SVD where we keep only $s$ of the highest singular values (in practice we use $s=32$; light transport is known to be low dimensional, in the sense that it loses detail similarly to a blur).  TODO

% A dual version of this construction could in principle be applied in generating the hidden video $L$ as well. However, we found that if both networks use this scheme, the optimization would quickly converge to a spatially inconsistent scrambled solution. We hypothesize that the stricter locality imposed by the direct convolutional filters in $\mathcal{L}$ is helpful and cannot be dropped entirely.

\paragraph{Implementation Details} The optimization is run using Adam~\cite{kingma2014adam} algorithm, simultaneously optimizing over parameters of $\mathcal{Q}$ and $\mathcal{L}$. %In these experiments, we set also the initial random noises $N_Q$ and $N_L$ as learnable optimization variables. 
The loss function is a sum of pointwise difference and a heavily weighted temporal gradient difference between $Z$ and reconstructed $TL$. Details are in the supplemental appendix. We extend the method to color by effectively treating it as three separate problems for R, G, and B; however, they become closely tied as the channels are generated by the same neural network as 3-dimensional output. We also penalize color saturation in the transport matrix to encourage the network to explain colors with the hidden video. To ensure non-negativity, we use a combination of exponentiations and tanh functions as output activations for the network $\mathcal{L}$. For $T$ we penalize negative values with a prior. We also found it useful to inject pixel and time coordinates as auxiliary feature maps, and to multiply a Hann window function onto intermediate feature maps at certain intermediate layers. These introduce forced spatial variation into the feature maps and help the networks to rapidly break symmetry in early iterations.

%We ensure non-negativity of $T$ by an additional prior that penalizes non-negative values in it. In case of $L$, we output two images and pass them to an exponential and a tanh function respectively, and sum the result. This encourages the network to maintain a constant intensity level, and adding and subtracting image content locally. We apply a low-weight smoothness prior on the response image dimension $IJ$ of $T$. We also anchor the average value of these slices to be the average value of the input video by a prior.

%To extend the method to color, we effectively treat it as three separate problems for R, G and B solved simultaneously. However, these are strongly bound together as they are output from the same network, simply by changing the output feature channel count. We also impose a prior that penalizes chromaticity in $T$, encouraging the optimization to explain color variations with the hiddeen video instead.

%entiating the output of the network.
%something about e.g. the coorndinate features in layers, exp/tanh in the end, extension to color, loss, priors, upsampling kernels, etc

\subsection{Experiments and Results}

\begin{figure*}[tb]
\makebox[\textwidth][c]{\includegraphics[width=\textwidth]{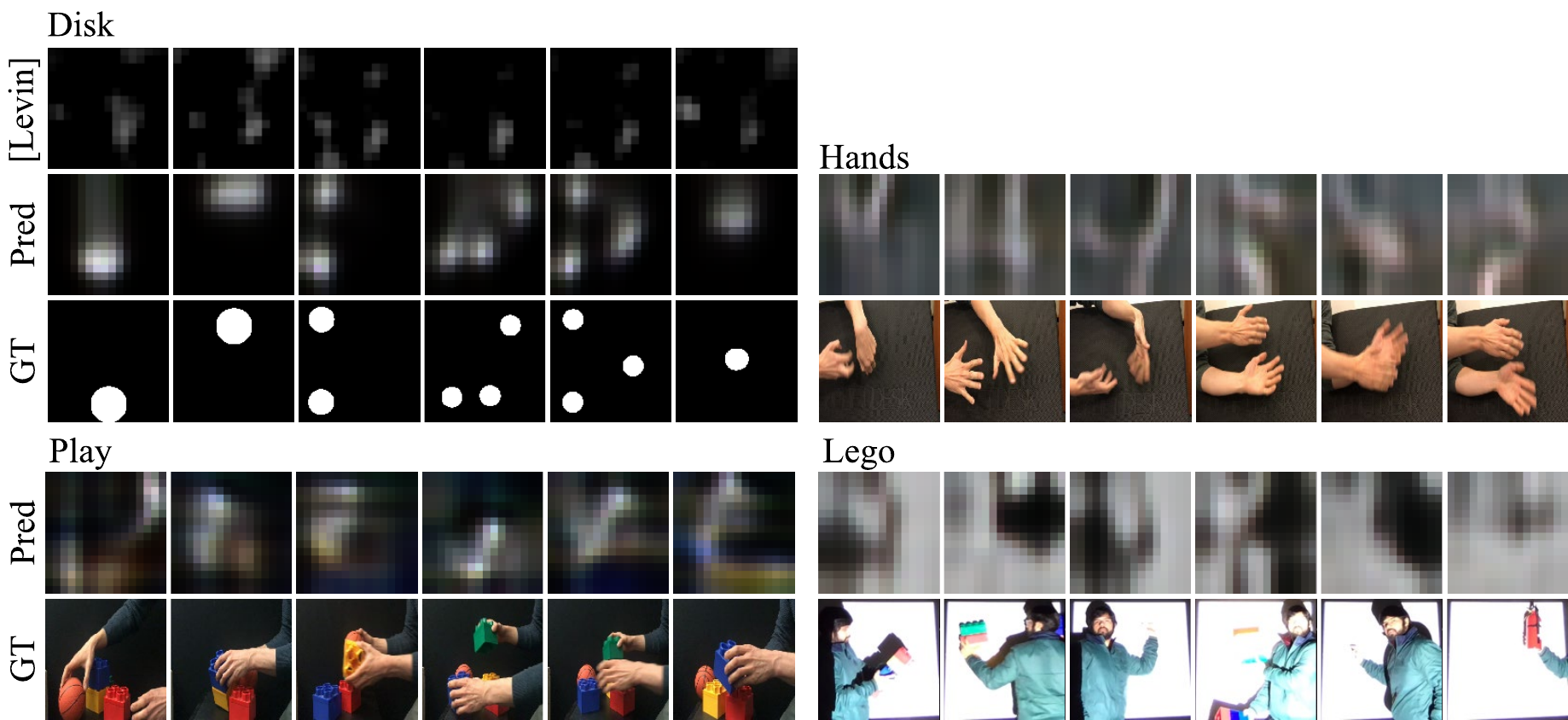}}
\caption{
Blind light transport factorization using our method. The first three sequences are projected onto a wall behind the camera. The \emph{Lego} sequence is performed live in front of the illuminated wall.}
\label{fig:results}
\end{figure*}

%\paragraph{Synthetic Data}
%\MA{Me and Lukas discussed it and decided that the synth data story is an unnecessary complication to the paper. Basically it works just the same as the real, and it's complicated to generate and explain, and nothing is learned by doing it, at least in what we did so far. So we decided to tentatively drop it. Agree?} \AY{Sounds good to me.}
%\PS{Are you planning to put it in supplementary? It can be helpful in making the point that the linearity assumption actually holds true since synth is linear for sure, and we get almost the same result with real capture. It's showing it by doing an explicit example. I do agree with the point that i } 

We test our method with multiple video datasets collected using a projector setup (as described in Section \ref{sec:inverselt}) recorded in different scenes with different hidden projected videos (Figure \ref{fig:results}).
%summarizes a selection of results with still frames from the extracted hidden video matrices. 
We encourage the reader to view the supplemental video, as motion is the main focus of this work. 
%We also visualize the recovered light transport matrices in the video.

The results demonstrate that our method is capable of disentangling the light transport from the content of the hidden video to produce a readable estimate of the latter. The \emph{disk} dataset is a controlled video showing variously complex motions of multiple bright spots. The number of the disks and their related positions are  resolved correctly, up to a spatial warp ambiguity similar to the one discussed in Section \ref{sec:distortions}. The space of ambiguities in this full 2-dimensional scenario is significantly larger than in the 1-D factorization: the videos can be arbitrarily rotated, flipped, shifted and often exhibit some degree of non-linear warping. The color balance between the factors is also ambiguous. As a control for possible unforeseen nonlinearities in the experimental imaging pipeline, we also tested the method on a semi-synthetic dataset that was generated by explicitly multiplying a measured light transport matrix with the \emph{disk} video; the results from this synthetic experiment were essentially identical to our experimental results.

The other hidden videos in our test set exhibit various degrees of complexity. For example, in \emph{hands}, we wave a pair of hands back and forth; watching our solved video, the motions and hand gestures are clearly recognizable. However, as the scenes become more complex, such as in a long fast-forwarded video showing colored blocks being variously manipulated (\emph{play}), the recovered video does show similarly colored elements moving in correlation with the ground truth, but the overall action is less intelligible.

%---such as a live-action video of a person manipulating lego blocks---the images become less intelligible. 
%Nevertheless, the position of the person and movements of his hands are often TODO, which is impressive, because the live-action nature of the video violates some our model's assumptions.
We also test our method on the live-action sequence introduced in Section \ref{sec:inverselt}. Note that in this scenario the projector plays no role, other than acting as a lamp illuminating the scene. While less clear than the baseline solution with a measured transport matrix, our blindly factored solution does still resolve the large-scale movements of the person, including movements of limbs as he waves his hands and rotates the Lego blocks.

%\NEW{The space of possible distortions in the full problem larger than in the one-dimensional case discussed in Section \ref{sec:distortions}. The ``benign'' class of distortions now include e.g. arbitrary complementary spatial rotations of the factors, and more complex continuous image warps. There is also a fundamental ambiguity about overall color of the factors.}

%\todo{what will we do with the half-finished distortion analysis thing?}

%\paragraph{Real Data with Live Action}

\paragraph{Comparison with Existing Approaches} 
%\todo{all very interesting but let's try to cut at least 30\% off this. Maybe the detailed description can be moved to appendix, and we could just very briefly summarize what was done. Also todo characterize the quality of the result in words a bit}

%Although there has been plenty of past work studying nonnegative matrix factorization~\cite{lee1999learning, virtanen2007monaural, hoyer2004non, salakhutdinov2008bayesian}, there has been to our knowledge none that studies it in a 4D light transport matrix recovery context. 

We compare our method to an extension of the deblurring approach by Levin et al. in~\cite{levin2011efficient}. We believe that blind deconvolution is the closest problem to ours, since it can be seen as a matrix factorization between a convolution matrix and a latent sharp image. We extended their marginalization method to handle general matrices and not just convolution, and use the same sparse derivative prior as them (see the supplementary materials for more details on how we adapted the approach). Figure~\ref{fig:results} and the supplementary video show that this approach produces vastly inferior reconstructions. %Our video results demonstrate a clear improvement over the method of~\cite{levin2011efficient}.

\section{Discussion and Conclusions}
% \vspace{-5pt}
% Because our algorithm assumes the light transport matrix to be linear, there are some inherent limitations to our approach. Moving scenes generally behave linearly, but one common source of non-linear behavior is self-occlusion within the moving scene. Sufficiently complex scenes that include self-occlusion could potentially pose problems for our method.

We have shown that cluttered scenes can be computationally turned into low-resolution mirrors without prior calibration. 
Given a single input video of the visible scene, we can recover a latent video of the hidden scene as well as a light transport matrix. 
We have expressed the problem as a factorization of the input video into a transport matrix and a lighting video, and used a deep prior consisting of convolutional neural networks trained in a one-off fashion. 
We find it remarkable that merely asking for latent factors easily expressible by a CNN is sufficient to solve our problem, allowing us to entirely bypass challenges such as the estimation of the geometry and reflectance properties of the scene. 

%Our assumption of linearity of light transport holds with some limitations in many realistic situations beyond the projector setup. Light transport is also perfectly linear with respect to the image of the hidden environment when it is far from the observed scene and no parallax occurs. In practice these conditions need not apply perfectly, as demonstrated e.g. by our live action experiment, where the person is not a projected image, and moves at a finite distance from the scene.

%There is a close relationship between our problem statement of matrix factorization and the problem of blind deconvolution; the latter is a special case with the ``transport'' matrix constrained to a Toeplitz-like structure. Convolutional models have been used before in to model NLoS imaging systems~\cite{yedidia2018analysis}. The goals of these approaches are different, as our primary aim is to merely untangle the scrambled hidden image into a visually readable form. Our results suggest that applying the Deep Image Prior to blind deconvolution could be a fruitful avenue of future work.

%We believe that the Deep Image Prior of Ulyanov et al.~\cite{ulyanov2018deep} can be fruitfully applied to a wide variety of different problems, as our results suggest. We are very excited about using their method in other domains as well, both because of its flexibility and because it does not require huge datasets to use. We believe that the Deep Image Prior represents a very important step forward in machine vision as a whole.

Blind inverse light transport is an instance of a more general pattern, where the latent variables of interest (the video) are tangled with another set of latent variables (the light transport), and to get one, we must simultaneously estimate both~\cite{Koenderink1997}. Our approach shows that when applicable, identifying and enforcing natural image structure in both terms is a powerful tool. 
We hope that our method can inspire novel approaches to a wide range of other apparently hopelessly ill-posed problems.

%\todo{Someone think of something uplifting and exciting to say in conclusion}

\subsubsection*{Acknowledgements}

This work was supported, in part, by DARPA under Contract No. HR0011-16-C-0030, and by NSF under Grant No. CCF-1816209. The authors wish to thank Luke Anderson for proofreading and helping with the manuscript.

%\NEW{TODO, DARPA at least?}           

% \clearpage
\bibliographystyle{plain}
{\small
\bibliography{factorization}
}

\section{Appendix: Implementation Details}

In the following, we discuss the technical implementation of the networks and losses presented in the paper. Please refer to the associated code release for the exact implementation details.

\subsection{Network Architectures}

The detailed architecture of the three networks used is discussed in Table \ref{fig:arch1} (the 1D matrix factorization task),  Table \ref{fig:arch2} (the video prediction network $\mathcal{L}$) and  Table \ref{fig:arch3} (the light transport singular vector weight prediction network $\mathcal{Q}$). Each network consists of a linear chain of layers, alternating between convolutions and upsampling. 
In particular, there are no downsampling cycles or skip connections.

\subsection{Loss and Training Details for Section 5.1}

The data fit loss the full blind inverse light transport problem is a sum of the following:

\begin{itemize}
\item Direct data fit loss $0.01 ~ ||TL - Z||_2^2$
\item Temporal gradient fit loss $||\nabla_t(TL - Z)||_1$ where $\nabla_t$ evaluates the (unnormalized) finite difference across dimension $t$. The finite difference interval is chosen at random on each iteration in $[1,8]$, so as to consider multiple time scales.
\end{itemize}

Priors are added to this loss:

\begin{itemize}
\item Nonnegativity prior $10 ~ ||\mathrm{min}(T,0)||_2$ over the transport matrix entries.
\item Smoothness prior $0.001 ~ ||\nabla_{IJ} T||_1$ over the observed image dimensions in the transport matrix $T$.
\item Color saturation prior $0.001 ||T - \mathrm{mean}_c(T)||_1$ that penalizes the difference of the transport matrix RGB channels from their average.
\item Magnitude prior $0.0001 ~ ||Q_0||_1$ where $Q_0$ are the weights for the first singular vector, i.e. the residual amount added on top of the mean. This prior merely anchors the magnitudes of the solution loosely, as $T$ and $L$ could accumulate arbitrary reciprocal multipliers over the iterations.
\end{itemize}

During the data loading phase, we identify any pixels that are overexposed (saturated) in any of the input frames, and exclude these pixels from all subsequent computations.

The colors are handled by predicting three output channels (R, G and B) in each network, and performing the matrix multiplication separately in each color channel. The basis singular vectors $U$ are also evaluated separately for each color channel. While in theory this would allow for mutually unrelated solutions for the three channels, in practice the color features naturally fall into coinciding positions, as they are built from common internal network features.

We train both networks simultaneously using the Adam optimizer~\cite{kingma2014adam} with learning rate $0.00006$.

We typically run the network for 100~000 iterations, which takes approximately four hours on an NVIDIA Titan Xp GPU. Typically we see coarse results at a few thousand iterations, and details become filled in over the remaining iterations. We implemented our model using PyTorch \cite{paszke2017automatic}.

\section{Comparison to Levin et al.~\cite{levin2011efficient}}

In Section 5.2 we compared the results achieved by our method to that achieved by an adapted version of the approach used by Levin et al. in~\cite{levin2011efficient}. We believe that this represents the closest existing analogue to a solution to our problem, since blind deconvolution can be seen as equivalent to matrix factoring when the light transport matrix must be Toeplitz, and in the problem of~\cite{levin2011efficient} both the base image and the blur kernel must exhibit spatial smoothness properties.

Following the example of~\cite{levin2011efficient}, we used an Expectation-Maximization (EM) framework to perform the joint reconstruction of the light transport matrix and the scene movie. In the E-step we solve for the mean light transport matrix given the scene movie, alone with an estimate of the associated covariance matrix over entries of the light transport matrix (for computational efficiency's sake, we assumed this covariance matrix was diagonal). In the M-step we solve for the best scene movie given the distribution over the light transport matrices implied by the mean and variance recovered in the E-step. In both steps, we penalized high spatial variations. We randomly initialized the light transport matrix and the scene movie, and iterated until the result converged.

\clearpage
\begin{table}[tbh]

% D:\ml\deblur\test\sequences\books\results\rdn\crop2
% \begin{tabular*}{20cm}{llllll}
\begin{tabularx}{\textwidth}{XXXXX}
 \textbf{id} & \textbf{type} & \textbf{output features} & \textbf{filter size}  & \textbf{activation} \\
\\
1  & conv & 32 & $4 \times 4$ & tanh \\
2 & \multicolumn{4}{l}{upsample} \\
3  & conv & 64 & $4 \times 4$ & tanh \\
4 & \multicolumn{4}{l}{upsample} \\
5  & conv & 64 & $4 \times 4$ & tanh \\
6 & \multicolumn{4}{l}{upsample} \\
7  & conv & 128 & $4 \times 4$ & tanh \\
8 & \multicolumn{4}{l}{upsample} \\
9  & conv & 128 & $4 \times 4$ & tanh \\
10 & \multicolumn{4}{l}{upsample} \\
11  & conv & 64 & $3 \times 3$ & leaky relu \\
12  & conv & 1 & $3 \times 3$ & exp \\
\end{tabularx}
\vspace{0.5cm}
\caption{Network architecture for $\mathcal{T}$ and $\mathcal{L}$ in Section 4.3. (plain matrix factorization, or ``1D light transport''). Both networks are essentially identical in architecture. The input to the network is a random 64-channel normal distributed tensor with $1/32$'th the spatial dimension of the target matrix. We include it as a learnable parameter of the optimization. The upsampling layers use bilinear interpolation. On every layer, we concatenate a pair of horizontal and vertical linear gradients (representing x and y coordinate values) as input feature maps. In the second-to-last layer, we concatenate as an auxiliary feature map the average of the input matrix rows. This corresponds to the average frame of the input ``video'', when that is the interpretation of the data. We also apply a pixel-wise dropout of probability $0.5$ on the input noise features.} 
\label{fig:arch1}
\end{table}

\begin{table}[tbh]

% D:\ml\deblur\test\sequences\books\results\rdn\crop2
% \begin{tabular*}{20cm}{llllll}
\begin{tabularx}{\textwidth}{XXXXX}
 \textbf{id} & \textbf{type} & \textbf{features} & \textbf{filter size}  & \textbf{activation} \\
\\
1  & conv & 64 & $3 \times 3 \times 3$ & leaky relu \\
2 & \multicolumn{4}{l}{upsample} \\
3  & conv & 64 & $3 \times 3 \times 3$ & leaky relu \\
4  & conv & 64 & $3 \times 3 \times 3$ & leaky relu \\
5 & \multicolumn{4}{l}{upsample} \\
6  & conv & 64 & $3 \times 3 \times 3$ & leaky relu \\
7  & conv & 64 & $3 \times 3 \times 3$ & leaky relu \\
8  & conv & 64 & $3 \times 3 \times 3$ & leaky relu \\
9 & \multicolumn{4}{l}{upsample} \\
10  & conv & 64 & $3 \times 3 \times 3$ & leaky relu \\
11  & conv & 32 & $3 \times 3 \times 3$ & leaky relu \\
12  & conv & 6 & $3 \times 3 \times 3$ & exp + tanh \\
\end{tabularx}
\vspace{0.5cm}
\caption{Network architecture for the hidden video network $\mathcal{L}$ in Section 5.1. The input to the network is a learnable random normal initialized tensor of 4 feature channels and $t/8 \times 2 \times 2$ (i.e. $2 \times 2$ pixels times $1/8$'th of the length of the video). The upsampling layers use nearest neighbor interpolation. Up to layer 7, all convolutional layers append three coordinate feature maps across space and time (with range $[-1,1]$). %In layers 4, 6 and 7 we also multiply a Hann window into the network outputs. 
The final layer outputs two RGB images -- one is passed to an exponential function and the other to a $b(\frac{1}{2} + \frac{1}{2}\mathrm{tanh}(x)$), where $b$ is a learnable RGB ``blacklevel'' parameter of dimension 3. The output of the tanh and exp are summed into the final predicted tensor $L$. The goal of the final activations is to prevent negative outputs by construction, and to bias the network towards using the learned black level as the ``default background color'' onto which it adds and subtracts image content. The output of the network is scaled by the reciprocal of the number of pixels $1/(16*16)$. The leaky relu slope parameter is $0.1$.}
\label{fig:arch2}
\end{table}

\begin{table}[tb]

% D:\ml\deblur\test\sequences\books\results\rdn\crop2
% \begin{tabular*}{20cm}{llllll}
\begin{tabularx}{\textwidth}{XXXXX}
 \textbf{id} & \textbf{type} & \textbf{features} & \textbf{filter size}  & \textbf{activation} \\
\\
1  & conv & 32 & $3 \times 3$ & leaky relu \\
2 & \multicolumn{4}{l}{upsample} \\
4  & conv & 64 & $3 \times 3$ & leaky relu \\
5  & conv & 64 & $3 \times 3$ & leaky relu \\
6  & conv & 64 & $3 \times 3 $ & leaky relu \\
7 & \multicolumn{4}{l}{upsample} \\
8  & conv & 64 & $3 \times 3 $ & leaky relu \\
9  & conv & 64 & $3 \times 3$ & leaky relu \\
10  & conv & 64 & $3 \times 3$ & leaky relu \\
11 & \multicolumn{4}{l}{upsample} \\
12  & conv & 128 & $3 \times 3 $ & leaky relu \\
13  & conv & 256 & $3 \times 3 $ & leaky relu \\
14  & conv & 96 & $3 \times 3$ & linear \\
\end{tabularx}
\vspace{0.5cm}
\caption{Network architecture for the singular value weight network $\mathcal{Q}$ used to build the light transport matrix in Section 5.1. The input to the network is a learnable randomly initialized tensor of 32 feature channels at spatial dimensions $2 \times 2$ pixels. The upsampling layers use nearest neighbor interpolation. Up to layer 10, all convolutional layers append coordinate feature maps across the two spatial dimensions. In layers 8, 9 and 10 we also multiply a Hann window into the network outputs. The final layer outputs 32 RGB images packed into 96 channels, one corresponding to each singular vector. After the final layer, these channels are multiplied by 96 learnable weights intended to provide the optimizer with direct means to control their strength. These are initialized as square roots of the singular values. After the multiplication by the singular vector basis (see Section 5.1), a precomputed mean image of the input video is added onto every slice of $T$, so that the network only needs to learn the residual.}
\label{fig:arch3}
\end{table}

%\clearpage
%\bibliographystyle{plain}
%{\small
%\bibliography{factorization}
%}

\end{document}